\documentclass[letterpaper]{article} 
\usepackage{aaai2027} 

\usepackage[hyphens]{url}  
\usepackage{graphicx} 
\usepackage{natbib}  
\usepackage{caption} 
\usepackage{algorithm}
\usepackage{algorithmic}
\usepackage{booktabs}
\usepackage{graphicx}
\usepackage{makecell}
\usepackage[table]{xcolor}
\usepackage{makecell}
\usepackage{amsmath}
\usepackage[table]{xcolor}
\usepackage{tabularx}
\usepackage{array}
\usepackage{newfloat}
\usepackage{listings}
\DeclareCaptionStyle{ruled}{labelfont=normalfont,labelsep=colon,strut=off} 
\floatstyle{ruled}
\newfloat{listing}{tb}{lst}{}
\floatname{listing}{Listing}
\usepackage[table]{xcolor}
\nocopyright 
\usepackage{booktabs}

\title{Overcoming Statistical Bias in Action-Controllable World Models}

\author{
Yuhong Shi\textsuperscript{\rm 1}, 
Zhenhao Chu \textsuperscript{\rm 1},  
Jie Wei \textsuperscript{\rm 2}, 
Jun Hao \textsuperscript{\rm 2}, 
Jianyi Liu \textsuperscript{\rm 1 *}, 
Jingwen Fu \textsuperscript{\rm 3 *}
    \\
}
\affiliations{
    \textsuperscript{\rm 1} Institute of Artificial Intelligence and Robotics, Xi'an Jiaotong University, Xi’an, China\\
    \textsuperscript{\rm 2} China Mobile (Shanghai) Information and Communication Technology Co., Ltd., Shanghai, China\\
    \textsuperscript{\rm 3} Zhongguancun Academy, Beijing, China\\

}

\begin{document}

\maketitle

\begin{abstract}
Action-conditioned world models aim to predict how visual environments evolve under an agent’s actions. Yet future frames are often highly predictable from visual inertia and recurring motion patterns alone. This creates a shortcut: models can fit the data by exploiting statistical biases without making their visible dynamics meaningfully depend on the action. As a result, different actions may produce similar futures, while motion may persist even under zero action. The key question is how to reduce reliance on statistical shortcuts from dominating action-conditioned prediction. We argue that action control requires more than injecting action features; it requires enforcing consistency under counterfactual changes to actions and observations. 
Based on this insight, we introduce \textbf{CoCo}, a Counterfactual Consistency framework to enhance action controllability through two complementary constraints. 
Multi-step counterfactual consistency constrains reference, inverse-action, and zero-action rollouts, while action-spatial counterfactual consistency enforces consistent predictions under mirrored scenes and transformed actions. Together, they reduce reliance on statistical shortcuts from substituting for action-dependent dynamics. We further introduce Action Response Consistency (ARC) and Drift Energy (DE) to assess action controllability, together with Mini-SSMB for same-state, multi-action counterfactual evaluation. 
On Mini-SSMB, our full model achieved ARC${_\mathrm{inv}}$ of 0.412 and ARC${_\mathrm{ref}}$ of 0.483, while reducing DE by 17.07\% relative to the baseline.
On VP$^2$ visual planning, it achieves the highest average success rate among SOTA models, at 73.1\%. Experiments on BAIR and RoboNet further show that these gains preserve video prediction quality and transfer across model settings.

\end{abstract}


\section{Introduction}

\begin{figure}[htbp]
  \centering
  \includegraphics[width=1\linewidth]{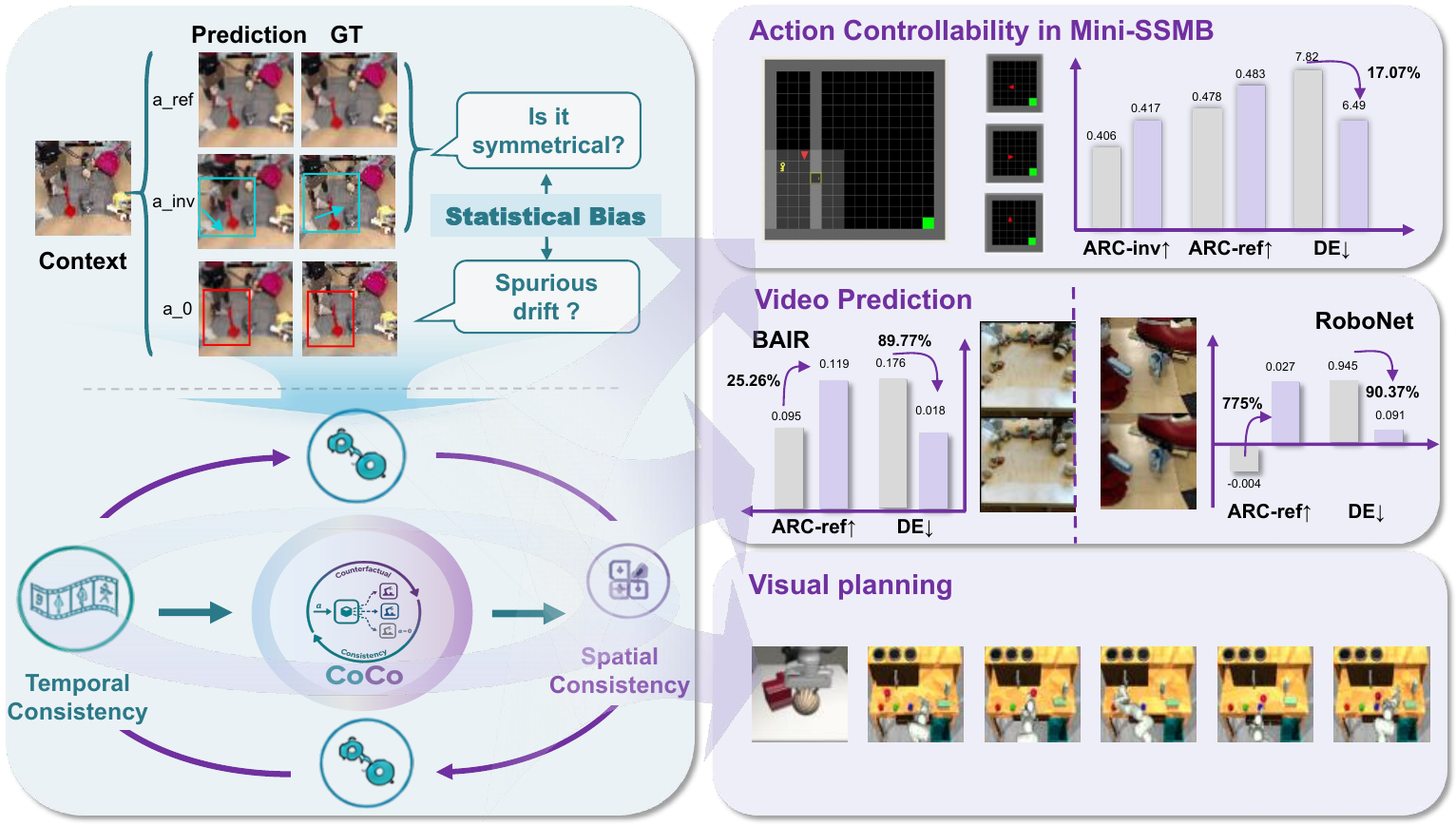}
  \caption{\textbf{Motivation and overview of CoCo.} 
  Existing models are affected by statistical biases, resulting in asymmetric responses between reference and inverse actions, as well as spurious drift in zero actions. CoCo mitigates these statistical shortcuts by simultaneously enforcing counterfactual consistency in both temporal and spatial dimensions.} 
  \label{fig1-1}
\end{figure}

Recent advances in autoregressive token models \cite{Wu2025RLVRWorld}, diffusion models \cite{Zhang2025DIMA}, and large-scale interactive simulators \cite{Bi2026Motus, Wu2025RLVRWorld} have substantially improved the fidelity and scale of visual world modeling.
At their core, world models are learned predictive simulators that enable agents to simulate future frames and evaluate candidate actions before interacting with the environment \cite{Zhou2025Persistent}.  
However, future frames are often highly predictable from visual inertia and recurring motion patterns alone, stemming from the statistical bias prevalent in data-driven learning, as shown in Fig. \ref{fig1-1}. 
When optimized to minimize empirical prediction error, a model naturally exploits the most frequent and easiest-to-predict correlations in the training data, rather than the mechanisms that actually determine future outcomes. This issue is particularly consequential for world models, since a planner relies on them to compare the consequences of alternative actions \cite{vafa2025inductive}. 
If distinct actions are mapped to nearly identical high-probability futures, the model may generate realistic videos while failing to function as an interactive simulator for decision-making \cite{li2025pinwm}.

Most existing work attempts to address this problem by strengthening how actions are represented and injected into world models \cite{Wang2024MetaDT}. Actions are incorporated through concatenated tokens \cite{2024iVideoGPT}, latent controls \cite{Maes2026LeWorldModel}, cross-attention \cite{Ma2024TransformerWorldModels}, feature modulation \cite{2024iVideoGPT}, or motion prompts \cite{wang2025sampo}, while complementary approaches improve representations and objectives through disentanglement and causal modeling \cite{lin2024because}, task-aware reconstruction \cite{Hutson2024Policy}, and robustness regularization \cite{ramasubramanian2025flat}. Although these methods can increase apparent action responsiveness and improve prediction accuracy on observed trajectories, they cannot remove statistical bias at its source \cite{pan2024counterfactual}. The reason is that their supervision remains observational. For each visual context, the training data contain only the future produced by the executed action, but not the futures that would have resulted from alternative or zero actions. Consequently, a model can satisfy the training objective by encoding the action while still predicting mainly from visual inertia, recurring motion, and dataset-level correlations \cite{gu2024denoising}. Standard factual losses cannot distinguish such shortcut predictions from genuinely action-driven transitions because both can match the future equally well \cite{liu2025fleet}. The limitation of prior work is therefore not insufficient action capacity, but the absence of supervision that identifies the causal effect of changing the action \cite{lancaster2024modemv2,yamada2024twist}. 
Without explicit counterfactual constraints, stronger action injection may reduce the symptoms of statistical bias, but cannot prevent the model from relying on it.

To address this issue, we propose \textbf{CoCo}, a {Co}unterfactual {Co}nsistency framework. 
CoCo is built on the principle that an action-driven world model should produce a correspondingly transformed future whenever an intervention changes the action, the scene, or both.
Therefore, temporal and spatial constraints are two complementary instances of the same structural requirement. 
In the temporal domain, an action sequence followed by its inverse should form a cycle that approximately recovers the initial state, whereas a zero action should act as an identity intervention that preserves the state. \textbf{Multi-Step Counterfactual Consistency (MSC2)} encodes these relations across rollouts, preserving the directional effect of actions. 
In the spatial domain, the joint transformation of the agent's current space and actions defines the equivariance relation of the world model, that is, the prediction of the mirror input should be equal to the predicted mirror.
\textbf{Action-Spatial Counterfactual Consistency (ASC2)} explicitly enforces this spatial equivariance. Together, these constraints enable CoCo to improve its measurement of the controllability of actions under the proposed counterfactual assessment, thereby enhancing its performance on downstream tasks.
Our contributions are summarized as:

\begin{itemize}
    \item We characterize how statistical shortcuts undermine action controllability in world models, showing that a model can achieve accurate factual prediction by exploiting visual regularities while failing to encode the observable effect of an action.
    
    \item We introduce counterfactual consistency as a structural principle for addressing statistical bias in action-conditioned world models. Under this principle, we propose MSC2 and ASC2, which jointly impose counterfactual constraints through temporal intervention relations and spatial transformation relations.
    
    \item We introduce Action Response Consistency (ARC) and Drift Energy (DE) to directly evaluate action-induced motion and zero-action stability, and construct Mini-SSMB for same-state, multi-action counterfactual evaluation. Extensive experiments validate the proposed framework on video prediction, visual planning, and model-based reinforcement learning.
\end{itemize}

\section{Related Work}

\subsection{Statistical Bias in World Models}

Recent studies show that predictive accuracy alone does not guarantee that a world model captures decision-relevant dynamics. Policy-shaped Prediction demonstrates that reconstruction-based models can allocate capacity to predictable yet policy-irrelevant visual distractions, thereby neglecting task-critical dynamics \cite{li2025lsimagine}. In MBRL \cite{hu2024imitation}, BECAUSE attributes the mismatch between accurate model prediction and policy performance to confounders in observational data, and mitigates it through causal state-action representations \cite{lin2024because}. Related work further addresses model bias and distribution shift during policy optimization \cite{feng2023finetuning,cheng2025jowa}, as well as robustness to optimization-induced model errors \cite{ramasubramanian2025flat}. More fundamentally, inductive-bias probes show that models can perform well on sequence prediction while relying on task-specific heuristics rather than learning the underlying dynamics \cite{vafa2025inductive}. These works reveal that world models can exploit statistical regularities instead of causal structure \cite{richens2024robust,gkountouras2025language,zhang2025creste}. 
Consequently, they leave an overlooked failure mode in which a world model can be accurate and robust yet remain action-insensitive at the pixel level, because it is never required to distinguish counterfactual futures induced by alternative actions from the same visual state.


\begin{figure*}[htb]
  \centering
  \includegraphics[width=0.9\linewidth]{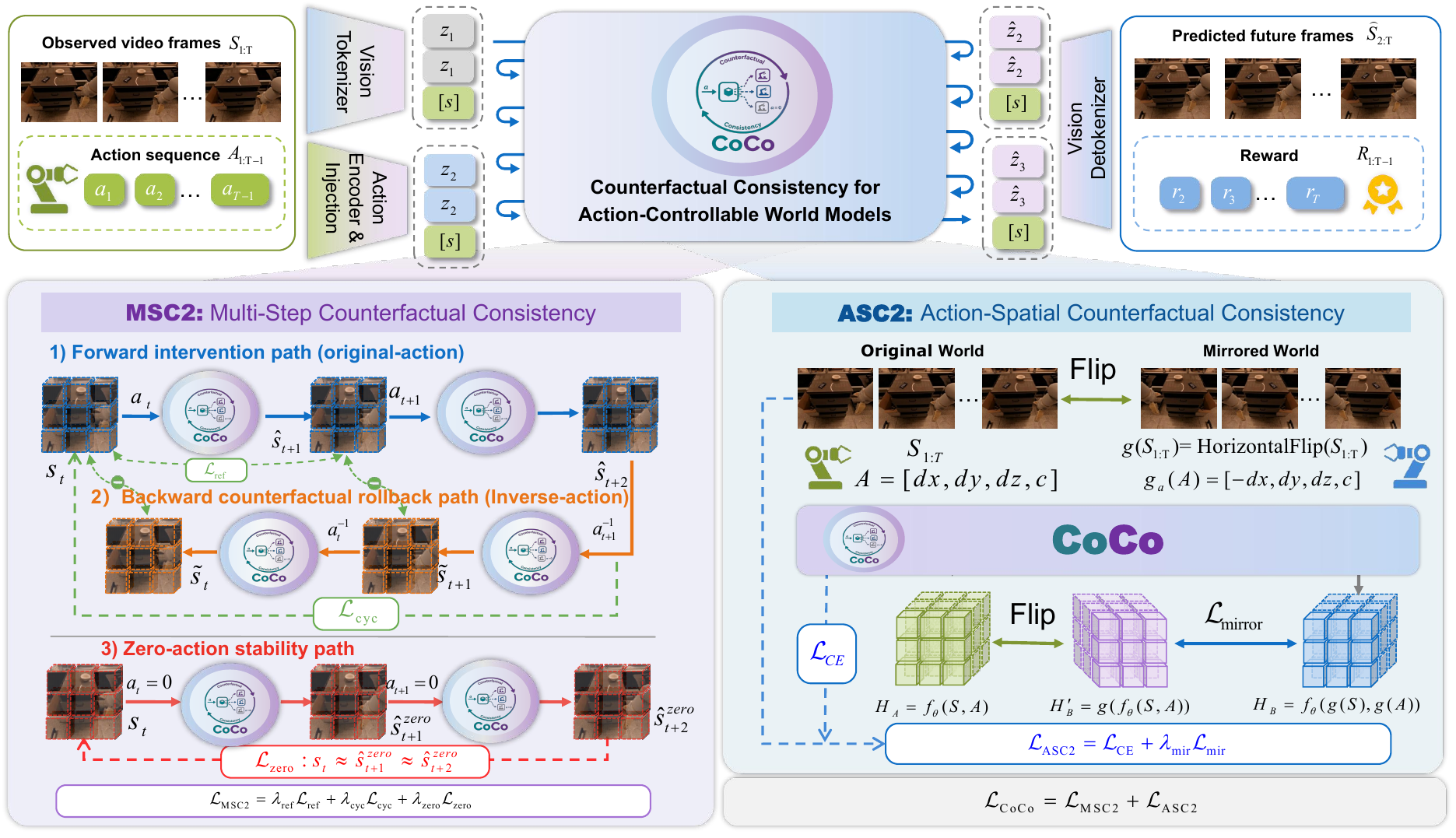}
  \caption{\textbf{Overview of the CoCo framework.}
Given observed frames and future actions, a shared action-conditioned Transformer predicts future video tokens, frames, and rewards. MSC2 improves temporal controllability by combining original-action forward prediction, inverse-action rollback, and zero-action stability constraints. ASC2 further enforces spatial action semantics by aligning predictions between the original and horizontally mirrored worlds with correspondingly transformed actions. }  \label{3-1}
\vspace{-1em}
\end{figure*}

\subsection{Counterfactual Consistency}

Counterfactual consistency asks whether a learned dynamics model predicts the appropriate change in future states after an intervention \cite{hao2025mosim, Bar_2025_CVPR,desousaribeiro2023high}. Adversarial counterfactual environment model learning identifies behavior-policy selection bias as a source of unreliable counterfactual prediction \cite{haugh2023counterfactual}, and uses adversarially weighted model learning to improve off-policy dynamics prediction and decision making \cite{chen2023adversarial,richens2024robust}. DALI \cite{roeder2025dali} learns latent environmental contexts from interactions and shows that intervening on context factors, such as gravity, induces physically plausible imagined rollouts. In visual modeling, CWM \cite{bear2023cwm} uses counterfactual input perturbations to expose structured physical factors, while CWMDT \cite{shen2025cwmdt} enables scene-level hypothetical interventions through digital-twin representations. Related controllable world models improve fine-grained motion prediction or long-horizon rollout quality through richer control signals and architectures \cite{jiang2025roboexp,hassan2025gem}.

\section{Method}

\subsection{Problem Definition}

Given an observed video sequence \(S_{1:T}=\{s_1,\ldots,s_T\}\) and its corresponding action sequence \(A_{1:T-1}=\{a_1,\ldots,a_{T-1}\}\), the objective of an action-conditioned world model is to learn a conditional generative mapping \(f_\theta\) that autoregressively predicts the next \(H\) frames from the historical context \(S_{\leq t}\) as
\begin{equation}
\hat{S}_{t+1:t+H}=f_\theta\!\left(S_{\leq t},A_{t:t+H-1}\right).
\label{e1}
\end{equation}
For models operating on discrete video tokens, the input frames are first encoded into a sequence of discrete latent variables by a tokenizer $Z_{1:T}=\mathcal{E}(S_{1:T}).$
An autoregressive Transformer \cite{Vaswani2017Attention} then models the conditional distribution
$p_\theta\!\left(Z_{t+1:t+H}\mid Z_{\leq t},A_{t:t+H-1}\right)$, 
and the predicted tokens are mapped back to pixel space through a detokenizer
$\hat{S}_{t+1:t+H}=\mathcal{D}\!\left(\hat{Z}_{t+1:t+H}\right)$.
Existing world models are typically trained using maximum likelihood estimation or a token-level reconstruction objective. 
%
%

However, this objective primarily encourages the model to generate plausible future frames under the observational data distribution. It does not ensure that the generated outcomes are genuinely controlled by the specified actions. In other words, the model generally learns the observational distribution
$p_\theta\!\left(Z_{t+1:t+H}\mid Z_{\leq t}, A_{t:t+H-1}\right)$,
rather than the interventional distribution of primary interest for robotic world modeling $p_{\theta}\bigl(Z_{t+1:t+H} \,\big|\, Z_{\leq t},\, \operatorname{do}\bigl(A_{t:t+H-1}\bigr)\bigr)$.
The difference lies in that the former allows the model to use historical visual inertia, background relevance, or statistical bias to predict the future, while the latter requires action to be the causal variable driving the change. Therefore, the focus of this paper is not merely on improving the quality of video reconstruction, but on overcoming this statistical bias and improving the controllability of action on the generated results.


\subsection{Multi-Step Counterfactual Consistency}


To address the aforementioned issues, we design a supervisory signal using counterfactual consistency in the world model prediction process, proposing the MSC2 module as shown in Fig. \ref{3-1}.
It aims to constrain the model to learn the causal relationship between actions and state changes by utilizing the multi-step consistency of the action world model and counterfactual action branches (i.e., the inverse action branch, and the zero action branch).
In principle, the original actions should drive consistent state transitions, the inverse actions should produce opposite changes, and the zero actions should suppress unrelated drift. 

Specifically, given the current state \(S_t\) and an action sequence
\(A_{t:t+K-1}=\{a_t,\cdots,a_{t+K-1}\}\), MSC2 evaluates the model under counterfactual rollouts. The original-action branch predicts the future $\hat{S}^{\mathrm{ref}}_{t+1:t+K} = f_\theta(S_t, A_{t:t+K-1})$,
where $H$ denotes the global video prediction horizon and satisfies $K\leq H$. MSC2 is constrained to follow the real action-induced dynamics by a multi-step prediction loss $\mathcal{L}_{\mathrm{ref}}
= \sum_{k=1}^{K} d\left(\hat{S}^{\mathrm{ref}}_{t+k}, S_{t+k}\right).$

To test whether the learned transition is action-reversible, we further construct an inverse-action sequence \(\mathcal{I}(A_{t:t+K-1})\), where each action is replaced by its inverse counterpart. Starting from the predicted endpoint \(\hat{S}^{\mathrm{ref}}_{t+K}\), the model performs a backward counterfactual rollout
\begin{equation}
\tilde{S}_{t}=f_\theta\left(\hat{S}^{\mathrm{ref}}_{t+K},
\mathcal{I}(A_{t:t+K-1})\right).
\end{equation}
If the model has learned action-consistent dynamics, the inverse rollout should recover the initial state, yielding the multi-step cycle constraint
$\mathcal{L}_{\mathrm{cyc}}=d\left(\tilde{S}_{t}, S_t\right)$. The forward reference rollout is greedily generated with stop-gradient. Its predicted endpoint is re-tokenized as the inverse-branch context, and the inverse rollout is optimized with teacher forcing.
In addition, MSC2 introduces a zero-action branch
\begin{equation}
\hat{S}^{\mathrm{zero}}_{t+1:t+K}=f_\theta(S_t, \mathbf{0}_{1:K}),
\end{equation}
which penalizes state drift when zero action is applied
$\mathcal{L}_{\mathrm{zero}}=\sum_{k=1}^{K}d\left(\hat{S}^{\mathrm{zero}}_{t+k}, S_t\right)$, where \(d(\cdot,\cdot)\) denotes the frame-wise mean absolute error.

Together, the three branches impose complementary constraints. The original-action branch aligns predicted motion with real dynamics, the inverse-action branch enforces reversibility under counterfactual inverse controls, and the zero-action branch suppresses action-independent drift. The resulting objective can be written as
\begin{equation}
\mathcal{L}_{\mathrm{MSC2}}=\lambda_{\mathrm{ref}}\mathcal{L}_{\mathrm{ref}}
+\lambda_{\mathrm{cyc}}\mathcal{L}_{\mathrm{cyc}}+\lambda_{\mathrm{zero}}\mathcal{L}_{\mathrm{zero}},
\end{equation}
where the non-negative coefficients $\lambda_{\mathrm{ref}}$,
$\lambda_{\mathrm{cyc}}$, and $\lambda_{\mathrm{zero}}$ balance the three objectives.

By applying these constraints to \(K\)-step counterfactual rollouts, MSC2 regularizes the consistency of generated trajectories beyond the standard one-step prediction objective.
It jointly enforces forward consistency, backward recovery, and zero-action stability, thereby encouraging the model to learn the direction and magnitude of future frame changes with action.


\subsection{Action-Spatial Counterfactual Consistency}

Multi-step counterfactual consistency constrains action-conditioned dynamics over time but does not capture the spatial meaning of actions. We therefore introduce ASC2, requiring the model’s predictions to transform consistently when the visual scene and corresponding actions undergo the same spatial transformation.

Let the action-conditional world model be \eqref{e1}, when a reflection-compatible action coordinate is available, for spatial transformation \(g\), there exists a corresponding action-space transformation \(g_a\) such that
\begin{equation}
g\left({f}_{\theta}\left(S_{\leq t}, A_{t:t+K-1}\right)\right)\approx{f}_{\theta}\left(g(S_{\leq t}), g_a(A_{t:t+K-1})\right).
\end{equation}
This equation expresses the spatial equivariance of action-conditioned dynamics: applying a spatial transformation to an action-conditioned future prediction should be equivalent to predicting the future from the correspondingly transformed observation and action.
For ASC2, $g$ is horizontal image mirroring. In MiniGrid, $g_a$ swaps {left} and {right} while preserving {forward} and {zero}. We use this as a dataset-dependent approximate regularizer, rather than assuming an exact physical symmetry.

Specifically, we construct two parallel branches with shared parameters. The original branch takes \((S_{\leq t}, A_{t:t+K-1})\) as input, while the mirrored branch takes \((g(S_{\leq t}), g_a(A_{t:t+K-1}))\). The CoCo produces continuous hidden representations at the future token positions
\begin{equation}
\begin{aligned}
H^{\mathrm{fut}}
&= f_{\theta}\left(S_{\leq t}, A_{t:t+K-1}\right), \\
\tilde{H}^{\mathrm{fut}}
&= f_{\theta}\left(
g(S_{\leq t}), g_a(A_{t:t+K-1})
\right),
\end{aligned}
\end{equation}
where $H^{\mathrm{fut}}$ and $\tilde{H}^{\mathrm{fut}}$ denote the continuous hidden representations for the original and mirrored branches, respectively.
Since the discrete codebook indices produced by the video tokenizer do not have inherent geometric meanings, we do not directly impose the mirroring constraint on token IDs. Instead, we align the continuous hidden representations. After reshaping the hidden states of future dynamic tokens into spatial grids, we apply a flipping operator \(F_w\) along the width dimension of the original branch and define the mirror-equivariance loss as 
$\mathcal{L}_{\mathrm{mir}}=\ell\left(F_w(H^{\mathrm{fut}}),\tilde{H}^{\mathrm{fut}}
\right),$
where we use the SmoothL1 loss for \(\ell\). The final objective of ASC2 is
\begin{equation}
\mathcal{L}_{\mathrm{ASC2}}=\mathcal{L}_{\mathrm{CE}}+\lambda_{\mathrm{mir}}\mathcal{L}_{\mathrm{mir}},
\end{equation}
where \(\mathcal{L}_{\mathrm{CE}}\) maintains the autoregressive prediction quality in the original visual space, \(\mathcal{L}_{\mathrm{mir}}\) encourages the model to learn action-spatial counterfactual consistency, and $\lambda_{\mathrm{mir}}$ is a non-negative balanced coefficient.


\section{Experiment}
We conduct experiments to answer four questions. \textbf{Q1:} whether conventional reconstruction metrics adequately reflect action controllability; \textbf{Q2:} whether the proposed counterfactual constraints improve action response and suppress action-independent drift; \textbf{Q3:} whether these gains hold across discrete and continuous control settings without sacrificing visual quality; and \textbf{Q4:} whether improved controllability benefits downstream decision-making. 


\subsection{Action Controllability Metrics}

Conventional world-model metrics, including FVD \cite{ge2024content}, LPIPS\cite{zhang2018unreasonable}, PSNR \cite{keles2021psnr}, and SSIM \cite{wang2004image}, primarily assess visual quality and cannot determine whether generated videos are controlled by the input actions.
To answer \textbf{Q1}, we introduce two counterfactual metrics, ARC and DE.
They capture complementary aspects of action controllability. ARC measures whether generated motion follows the action, whereas DE measures whether the model remains stable in the zero action. 
By comparing predictions under reference, inverse, and zero actions, these metrics more directly assess action-driven dynamics than conventional visual-quality metrics. Detailed definitions are provided in the Appendix A.

\begin{figure}[htb]
  \centering
  \includegraphics[width=1\linewidth]{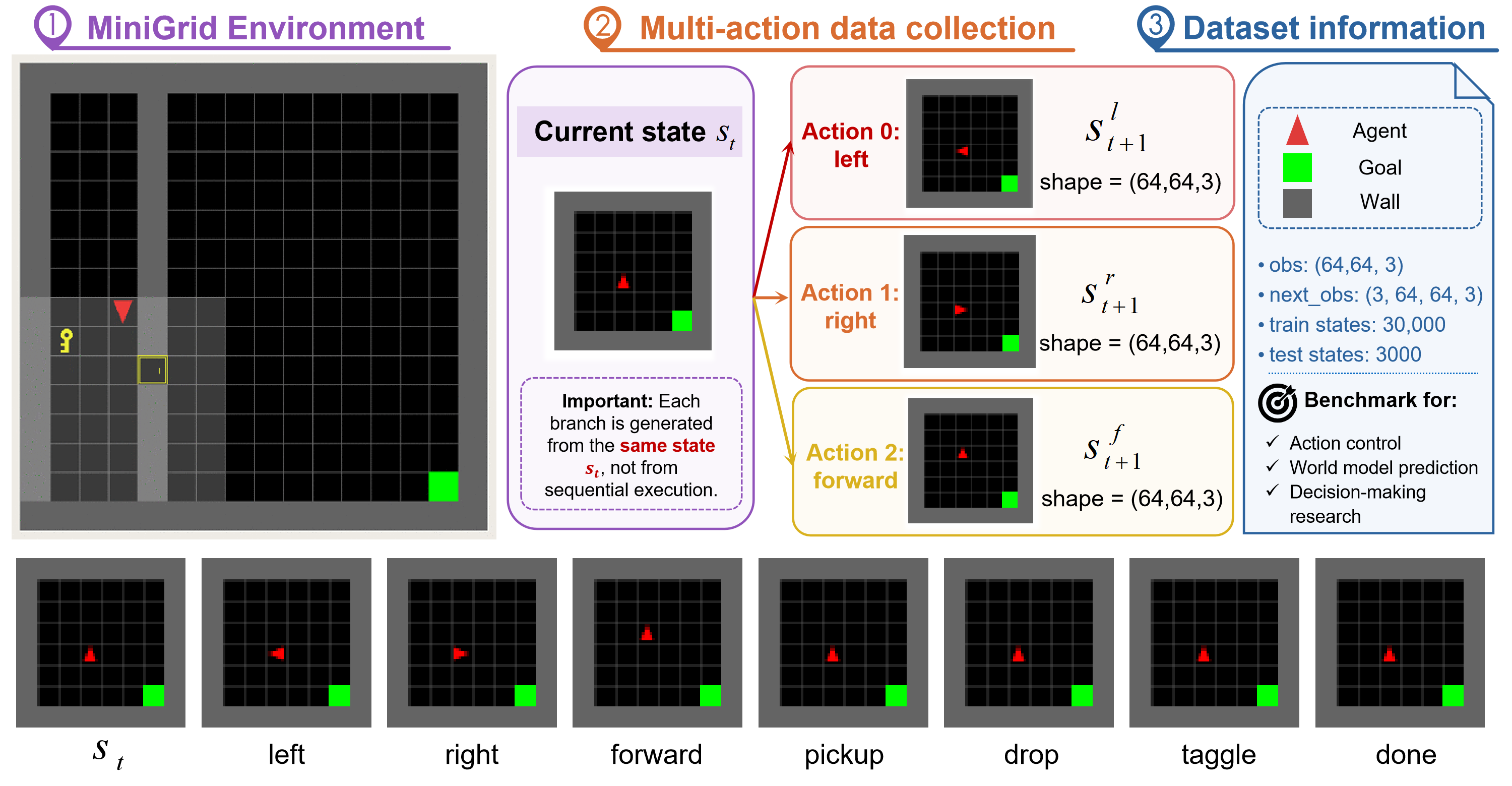}
  \caption{\textbf{The overview of the Mini-SSMB Dataset. }
  Mini-SSMB contains 30,000 training and 3,000 test states, each paired with next-frame observations for left, right, and forward. Left and right rotate the agent, while forward moves it ahead unless blocked by a wall. 
  }  \label{mini}
\vspace{-1em}
\end{figure}

\begin{figure}[htb]
  \centering
  \includegraphics[width=1\linewidth]{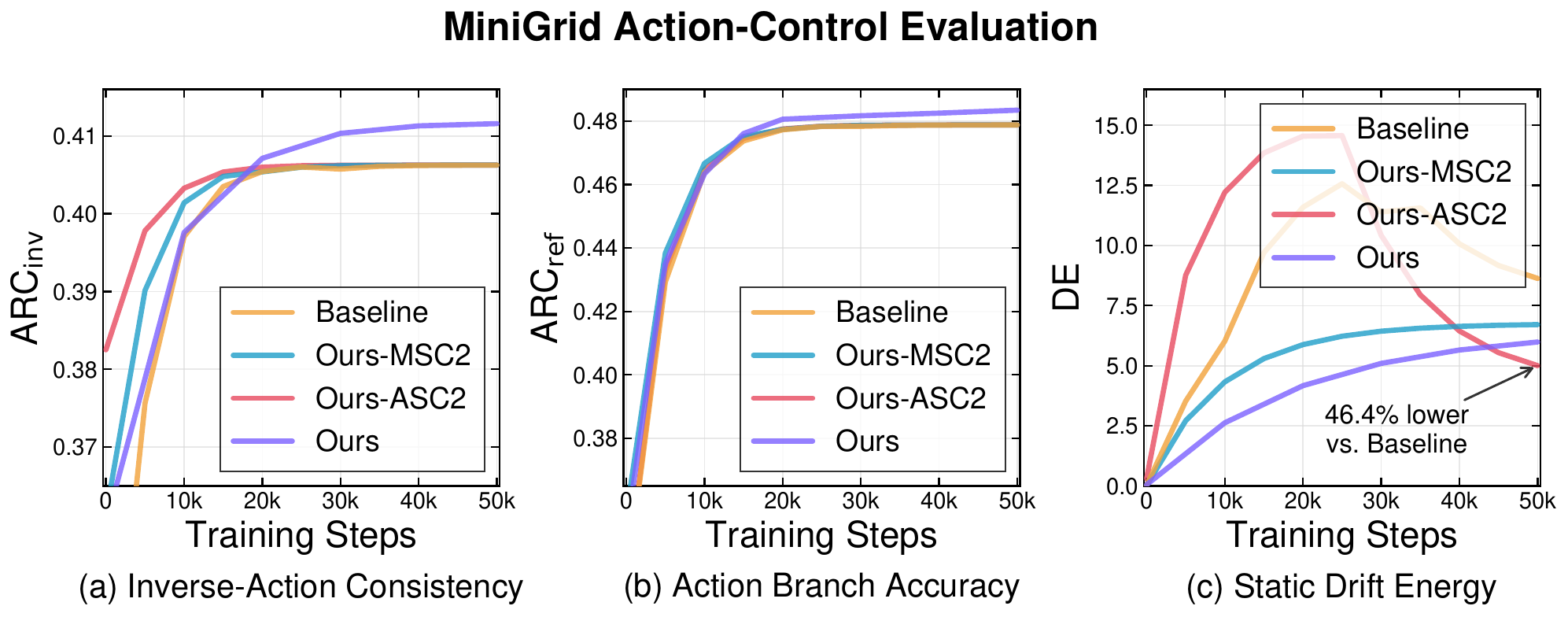}
  \caption{\textbf{Training dynamics of action-control metrics on Mini-SSMB.} 
Ours-MSC2 and Ours-ASC2 each contain one module, and Ours serves as the full model. The baseline model is presented by iVideoGPT.}\label{fig3}
\vspace{-1em}
\end{figure}

\subsection{Action Controllability in Mini-SSMB}

\subsubsection{Setup} 
\textit{1) Dataset construction:} To systematically evaluate CoCo's ability to controllably generate discrete actions (\textbf{Q2}), we constructed a dataset based on MiniGrid with the same state and multiple branches (Mini-SSMB). Unlike single-trajectory datasets, Mini-SSMB applies multiple actions to the same state \(s_t\) and records their corresponding next states \(s_{t+1}^{a}\). As in Fig. \ref{mini}, we use MiniGrid \cite{Cheva2023MinigridMiniworld} with 64\(\times\)64 RGB observations and three navigation actions: left, right, and forward. Detailed information about the Mini-SSMB is provided in Appendix B.



\textit{2) Training and evaluation.}
We train our model on Mini-SSMB to evaluate action-state consistency. All models are initialized from the pretrained checkpoint and trained with two-frame clips. We compare the baseline with Ours-MSC2, Ours-ASC2, and the full Ours model, using 50K optimization steps, batch size 16, learning rate (1$\times$10$^{-5}$), cosine scheduling, and (64$\times$64) resolution. All experiments are conducted on a server equipped with eight NVIDIA A800 GPUs.

\begin{table*}[htb]
\centering
\setlength{\tabcolsep}{3pt} 
\small
\begin{minipage}[t]{0.49\textwidth}
\vspace{0pt} \centering
\begin{tabularx}{\linewidth}{
>{\centering\arraybackslash}X
*{3}{>{\centering\arraybackslash}p{0.15\linewidth}}}

\toprule
Model
& ARC$_{\mathrm{inv}}\uparrow$
& ARC$_{\mathrm{ref}}\uparrow$
& $\mathrm{DE}\downarrow$ \\  \midrule
\multicolumn{4}{c}{\textit{Action-conditioned at 64${\times}$64 resolution on BAIR}} \\  \midrule
CA & 0.022& -0.022& 0.239 \\
MR          & 0.034& -0.034& 0.272 \\
FitVid      & -0.062& -0.302& 0.050 \\
iVideoGPT   & 0.074& 0.095& 0.176 \\
Ours-MSC2   & 0.083& 0.105& 0.023 \\
Ours-ASC2    & 0.086& 0.102& 0.071 \\
\rowcolor{gray!15}
\textbf{Ours}       & \textbf{0.100}& \textbf{0.119}& \textbf{0.018} \\  \midrule
\multicolumn{4}{c}{\textit{Action-conditioned at 64${\times}$64 resolution on RoboNet}} \\  \midrule
CA & 0.002& -0.002& 0.253 \\
MR  & 0.008& -0.007& 0.218 \\
FitVid  & -0.045& -0.158& 0.921 \\
iVideoGPT  & 0.013& -0.004& 0.945 \\
Ours-MSC2	& 0.049 &	0.036 &	0.095 \\
Ours-ASC2 &	0.042 &	0.027 &	0.104 \\
\rowcolor{gray!15}
\textbf{Ours} &	\textbf{0.055} &	\textbf{0.064} &	\textbf{0.091}  \\  
\bottomrule
\end{tabularx}  \end{minipage}  \hfill
\begin{minipage}[t]{0.49\textwidth}
\vspace{0pt} \centering
\begin{tabularx}{\linewidth}{
>{\centering\arraybackslash}X
*{4}{>{\centering\arraybackslash}p{0.125\linewidth}}}

\toprule Model
& FVD $\downarrow$
& LPIPS $\downarrow$
& PSNR $\uparrow$
& SSIM $\uparrow$ \\ \midrule
\multicolumn{5}{c}{\textit{Action-conditioned at 64${\times}$64 resolution on BAIR}} \\ \midrule
CA  & 382.48& 12.74& 18.29& 77.88 \\
MR  & 375.63& 12.75& 18.29& 77.88 \\
iVideoGPT  & 73.46& 5.88& 23.33& 88.17 \\
SAMPO & \textbf{55.50}& 5.00& 26.70& \textbf{94.70} \\
Ours-MSC2& 73.12& 4.92& 26.38& 88.27 \\
Ours-ASC2& 74.15& 4.87& 27.36& 88.18 \\
\rowcolor{gray!15}
\textbf{Ours}& 71.27& \textbf{4.84}& \textbf{28.35}& 88.10 \\  \midrule
\multicolumn{5}{c}{\textit{Action-conditioned at 64${\times}$64 resolution on RoboNet}} \\  \midrule
FitVid  & 62.50& 2.40& 28.20& 89.30 \\
MaskViT  & 133.50& 4.20& 23.20& 80.50 \\
iVideoGPT & 94.07& 2.01& 19.35& 65.87 \\
SAMPO & 57.10& 3.30& \textbf{29.30}& \textbf{94.10} \\
Ours-MSC2& 67.01& 1.57& 26.15& 82.78 \\
Ours-ASC2& 59.77& 1.36& 27.27& 85.60 \\
\rowcolor{gray!15}
\textbf{Ours}& \textbf{43.81}& \textbf{1.32}& 27.33& 85.75 \\  \bottomrule
\end{tabularx} \end{minipage}
\caption{\textbf{Video prediction and action controllability performance on the BAIR and RoboNet datasets.}
We evaluate action-conditioned world models on BAIR and RoboNet using both controllability metrics and standard visual quality metrics. Each reported metric is calculated as the mean and standard deviation over three runs. LPIPS and SSIM scores are scaled by 100 for convenient display.} \label{tab:video_prediction_results}
  \vspace{-1em}
\end{table*}

\begin{figure*}[htb]
  \centering
  \includegraphics[width=0.88\linewidth]{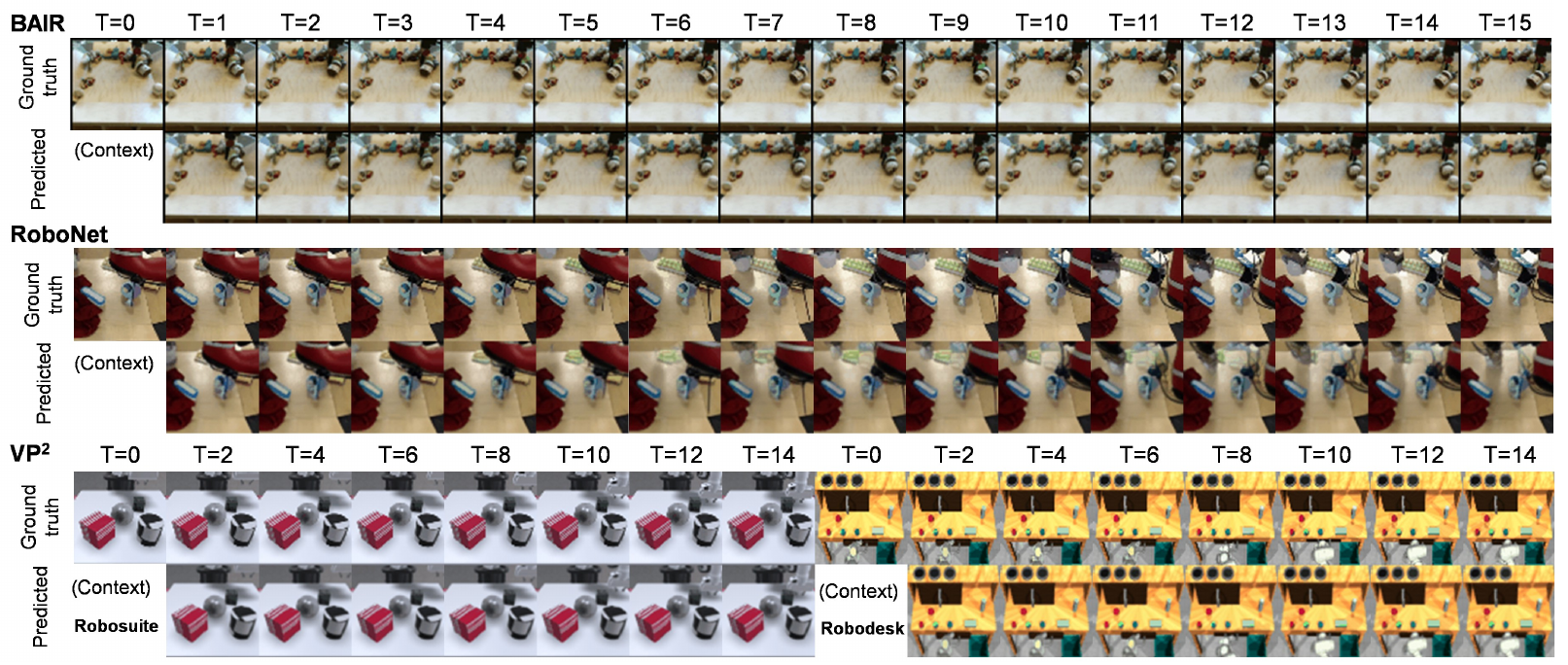}
  \caption{\textbf{Visualization of video prediction results across datasets.}
  We visualize ground-truth and predicted rollouts on BAIR, RoboNet datasets, and VP$^2$ tasks including robosuite and RoboDesk. 
  }  \label{fig4}
  \vspace{-1em}
\end{figure*}

\begin{figure*}[htb]
  \centering
  \includegraphics[width=0.86\linewidth]{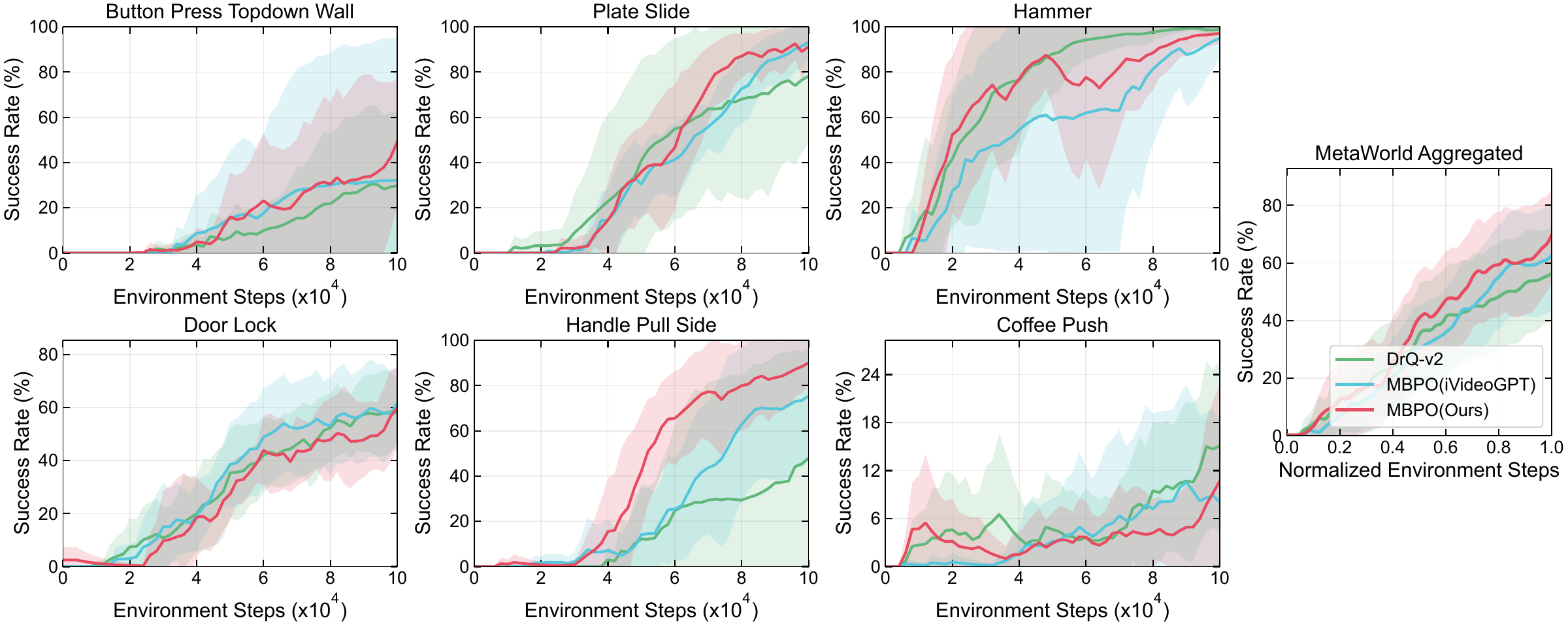}
  \caption{\textbf{Model-based reinforcement learning on MetaWorld.} Success rate learning curves on six manipulation tasks and their normalized aggregate. Curves show the mean over five random seeds, and shaded regions denote 95\% confidence intervals.}
  \label{f6}
    \vspace{-1em}
\end{figure*}

\subsubsection{Results Analysis}  
Fig. \ref{fig3} shows a comparison of the action controllability of our model and the baseline. All models achieve ARC scores above the random baseline of \(1/3\), indicating that they successfully learn the basic action-state correspondence. However, the base model still exhibits substantial drift under zero actions (see Fig.\ref{fig3} (c)).  
Ours-MSC2 model reduces DE from 7.82 to 6.75 (13.7\%), while Ours-ASC2 further lowers it to 4.19 (46.4\%), without sacrificing reference- or inverse-action accuracy. The higher motion ARC (0.624) than rotation ARC (0.406) also shows that forward displacement is easier to capture than fine-grained in-place rotations. Overall, CoCo achieves better control over predicted following actions and effectively suppresses action-independent variations, resulting in better dynamic control than baseline methods.

\begin{table*}[htb]
\centering
\setlength{\tabcolsep}{6pt}
\small
\resizebox{\textwidth}{!}{
\begin{tabular}{ccccccccc|c}
\toprule
Method / Task & \makecell{robosuite\\Push} & \makecell{Flat\\Block} & \makecell{Open\\Drawer} & \makecell{Open\\Slide} & \makecell{Blue\\Button} & \makecell{Green\\Button} & \makecell{Red\\Button} & \makecell{Upright\\Block} & \makecell{Avg.\\Success$\uparrow$} \\
\midrule
Simulator & 93.5$^{\scriptscriptstyle\pm 1.2}$ & 13.3$^{\scriptscriptstyle\pm 0.0}$ & 76.7$^{\scriptscriptstyle\pm 0.0}$ & 71.7$^{\scriptscriptstyle\pm 1.4}$ & 100.0$^{\scriptscriptstyle\pm 0.0}$ & 96.7$^{\scriptscriptstyle\pm 0.0}$ & 90.0$^{\scriptscriptstyle\pm 0.0}$ & 90.0$^{\scriptscriptstyle\pm 0.0}$ & 100.0 \\
FitVid   & 67.7$^{\scriptscriptstyle\pm 6.4}$ & 9.2$^{\scriptscriptstyle\pm 2.8}$ & 25.3$^{\scriptscriptstyle\pm 8.2}$ & 35.3$^{\scriptscriptstyle\pm 5.5}$ & 94.0$^{\scriptscriptstyle\pm 4.2}$ & 84.0$^{\scriptscriptstyle\pm 5.5}$ & 58.7$^{\scriptscriptstyle\pm 5.5}$ & 51.3$^{\scriptscriptstyle\pm 2.9}$ & 65.6 \\
MCVD   & 77.3$^{\scriptscriptstyle\pm 2.1}$ & 4.0$^{\scriptscriptstyle\pm 1.4}$ & 11.7$^{\scriptscriptstyle\pm 1.4}$ & 18.3$^{\scriptscriptstyle\pm 1.4}$ & 95.0$^{\scriptscriptstyle\pm 4.1}$ & 83.3$^{\scriptscriptstyle\pm 0.0}$ & 73.3$^{\scriptscriptstyle\pm 2.7}$ & 56.7$^{\scriptscriptstyle\pm 2.7}$ & 64.3 \\
MaskViT  & 82.6$^{\scriptscriptstyle\pm 2.5}$ & 4.0$^{\scriptscriptstyle\pm 4.1}$ & 4.0$^{\scriptscriptstyle\pm 4.1}$ & 8.7$^{\scriptscriptstyle\pm 5.5}$ & 94.7$^{\scriptscriptstyle\pm 1.4}$ & 64.0$^{\scriptscriptstyle\pm 4.1}$ & 24.0$^{\scriptscriptstyle\pm 8.2}$ & 62.2$^{\scriptscriptstyle\pm 9.5}$ & 52.1 \\
iVideoGPT   & 78.3$^{\scriptscriptstyle\pm 0.8}$ & 3.3$^{\scriptscriptstyle\pm 0.7}$ & 37.5$^{\scriptscriptstyle\pm 1.7}$ & 16.1$^{\scriptscriptstyle\pm 2.7}$ & 95.6$^{\scriptscriptstyle\pm 2.1}$ & 82.5$^{\scriptscriptstyle\pm 3.4}$ & 92.2$^{\scriptscriptstyle\pm 1.4}$ & 44.7$^{\scriptscriptstyle\pm 1.7}$ & 70.1 \\
SAMPO  & 80.7$^{\scriptscriptstyle\pm 1.4}$ & 5.5$^{\scriptscriptstyle\pm 1.2}$ & 40.3$^{\scriptscriptstyle\pm 2.3}$ & 18.3$^{\scriptscriptstyle\pm 3.3}$ & 97.3$^{\scriptscriptstyle\pm 1.7}$ & 85.3$^{\scriptscriptstyle\pm 3.3}$ & 94.7$^{\scriptscriptstyle\pm 2.1}$ & 46.1$^{\scriptscriptstyle\pm 2.7}$ & 72.2 \\
\midrule
\rowcolor{gray!15}
\textbf{Ours} & 83.8$^{\scriptscriptstyle\pm 1.9}$ & 4.1$^{\scriptscriptstyle\pm 2.4}$ & 32.5$^{\scriptscriptstyle\pm 7.9}$ & 21.5$^{\scriptscriptstyle\pm 4.3}$ & 98.3$^{\scriptscriptstyle\pm 1.7}$ & 88.3$^{\scriptscriptstyle\pm 3.7}$ & 89.8$^{\scriptscriptstyle\pm 2.8}$ & 54.2$^{\scriptscriptstyle\pm 4.3}$ & \textbf{73.1} \\
\bottomrule
\end{tabular}}
\caption{\textbf{Visual planning performance in VP$^2$.}
We report the success rates across 7 tasks in RoboDesk and 1 task in robosuite, and the average success rate is calculated by normalizing against the Simulator's success rate and excluding FlatBlock. In addition, we provide the mean and standard deviation of the success rates (in \%) on average across 4 random seeds.}\label{tab:visual_planning_vp2}
  \vspace{-1em}
\end{table*}

\subsection{Video Prediction}
\subsubsection{Setup}
To answer \textbf{Q3} and evaluate the action-conditional video prediction performance of CoCo, we conducted experiments on the BAIR \cite{Ebert2017BAIRRobotPushing} and RoboNet~\cite{Dasari2020RoboNet}. BAIR contains 64\(\times\)64 RGB videos of robotic pushing interactions with 4-D continuous actions, providing a controlled setting for evaluating whether the model can capture fine-grained action-induced object motion. RoboNet is a larger and more diverse multi-robot manipulation dataset; we use its preprocessed 64\(\times\)64 videos with 5-D continuous control signals to evaluate generalization under more heterogeneous robots, scenes, and object interactions.

\subsubsection{Results Analysis} 

The baselines for this experiment can be divided into two categories. The first includes action-enhanced methods, such as CA\cite{perez2018film} and MR\cite{zhang2023controlnet}, which inject action information through conditional modulation or motion residuals but do not explicitly enforce counterfactual controllability. 
The second includes video prediction or action-conditioned baselines, such as FitVid \cite{Babaeizadeh2021FitVid}, MaskViT \cite{Gupta2023MaskViT}, iVideoGPT \cite{2024iVideoGPT}, and SAMPO~\cite{wang2025sampo}, which mainly optimize visual prediction quality. 
%
As shown in Table \ref{tab:video_prediction_results}, on action controllability, CoCo consistently achieves the best results. On BAIR, the full Ours obtains the highest ARC$_{\mathrm{inv}}$ and ARC$_{\mathrm{ref}}$ scores, 0.100 and 0.119, outperforming iVideoGPT by 35.1\% and 25.3\%, respectively. It also reduces DE from 0.176 to 0.018, showing a substantial reduction in static drift. Ours-MSC2 already lowers DE to 0.023, confirming that counterfactual consistency directly improves zero-action stability, while Ours-ASC2 improves ARC$_{\mathrm{inv}}$ to 0.086, indicating better directional action response. 
The improvement is more pronounced on RoboNet, where existing baselines suffer from severe drift. iVideoGPT and FitVid obtain DE values of 0.945 and 0.921, respectively, while our model reduces DE to 0.091. Meanwhile, Ours achieves the best ARC$_{\mathrm{inv}}$ and ARC$_{\mathrm{ref}}$ scores among action-conditioned models, indicating that the generated transitions are better aligned with the conditioned action branch rather than visual inertia or spurious dynamics. 

Visualization of video predictions is shown in Fig. \ref{fig4}.
For visual prediction quality, Ours remains competitive and often improves over the action-conditioned backbone. On BAIR, Ours reduces FVD from 73.46 to 71.27, LPIPS from 5.88 to 4.84, and improves PSNR from 23.33 to 28.35 compared with iVideoGPT. On RoboNet, Ours achieves the best FVD of 43.81 and the best LPIPS of 1.32 among the compared models. 
Although SAMPO achieves higher PSNR (29.30) and SSIM (94.10), likely due to its scale-wise autoregressive design prioritizing pixel-level fidelity, CoCo demonstrates stronger action controllability by explicitly enforcing counterfactual consistency. Overall, these results indicate that CoCo effectively mitigates statistical bias while maintaining competitive video generation quality. 
Sensitivity analysis of hyperparameters are in Appendix C.

\subsection{Visual Planning}

\subsubsection{Setup}
To answer \textbf{Q4} and evaluate object-conditional visual planning, we conduct experiments on the VP$^2$ benchmark~\cite{tian2023vp2} using both RoboDesk and robosuite. 
The planner receives RGB observations resized to 64\(\times\)64, conditions the video model on two context frames, and predicts future observations under candidate action sequences. 
We evaluate RoboDesk tasks including object pushing, drawer/slide opening, and block-off-table manipulation, and robosuite manipulation tasks with the \texttt{agentview\_shift\_2} camera. 
RoboDesk uses a 5-D continuous action space, while robosuite uses a 4-D action space.
At each step, we use CEM to optimize candidate action sequences over a horizon of 10 by minimizing the RGB mean-squared error between predicted frames and the goal image. 
%

\subsubsection{Results Analysis}  
Table \ref{tab:visual_planning_vp2} compares three classes of methods: generic video prediction baselines (FitVid, MCVD, MaskViT, and iVideoGPT), an action-conditioned planning baseline (SAMPO), and our model.
Our method achieves the best average success rate 73.1, outperforming iVideoGPT by 3.0 points and SAMPO by 0.9 points. This improvement is especially clear on tasks that require stable directional execution and long-horizon control, such as robosuite Push, Open Slide, Blue Button, Green Button, and Upright Block, where our model consistently matches or exceeds the strongest baselines.
These gains are consistent with our design. The cycle constraint reduces rollout drift and enforces action reversibility, while the equivariance constraint aligns action direction with spatial action. As a result, the model converts action information into more reliable planning behavior rather than merely producing visually plausible rollouts. 


\subsection{Model-based Reinforcement Learning}

\subsubsection{Setup}
To answer \textbf{Q4}, we conducted experiments on six metaworld operation tasks to verify the performance of reinforcement learning based on the visual model. The agent receives RGB observations from the corner camera with a frame stack of 3, and acts in the continuous 4-D MetaWorld action space. Each episode lasts up to 100 environment steps. 
We compare the model-free baseline DrQ-v2 \cite{yarats2022mastering} with the model-based agent in MBPO \cite{2024iVideoGPT}, which is augmented by a pre-trained world model. We compare the performance of the CoCo model with the iVideoGPT model in MBPO. 


\subsubsection{Results Analysis}  


Fig.~\ref{f6} shows that CoCo improves the aggregate success rate of MBPO over both the iVideoGPT-based world model and the model-free DrQ-v2 baseline. 
The gain is most pronounced on Button Press Topdown Wall and Handle Pull Side, where MBPO(Ours) achieves faster learning and a substantially higher final success rate. 
On Plate Slide, CoCo reaches a similar final performance to MBPO(iVideoGPT) while improving learning progress in the later stage.
It remains competitive on Hammer and Door Lock, indicating that the counterfactual constraints do not compromise performance on tasks where the baseline world model is already effective.  
Yet Coffee Push remains challenging for all methods, and CoCo does not provide a clear advantage on this task. 
Overall, these results suggest that the proposed counterfactual constraints can benefit model-based policy learning, particularly for tasks requiring reliable action-conditioned state transitions. The overlapping confidence intervals on several tasks indicate task-dependent gains and motivate broader evaluation in future work.


\section{Conclusion}

This paper explores how to reduce the impact of statistical bias on the controllability of actions in world models through counterfactual consistency.
 We introduce CoCo, a counterfactual consistency framework that enforces multi-step intervention consistency and spatial equivariance through MSC2 and ASC2, respectively. Together with the ARC and DE metrics and the Mini-SSMB dataset, CoCo makes action-controllable world models trainable and measurable. Experiments on video prediction and downstream decision-making tasks show that CoCo improves action controllability while maintaining competitive video prediction quality.

However, constructing multiple counterfactual branches and spatially transformed inputs introduces additional computation. Moreover, predefined inverse actions and spatial transformations may be ill-defined for irreversible contacts or complex semantic actions. In the future, we will investigate learning valid transformations directly from data and improving the efficiency of counterfactual training.

\clearpage
\bibliography{aaai2027}


\end{document}